\pdfoutput=1

\documentclass[11pt]{article}

\usepackage{acl}

\usepackage{times}
\usepackage{latexsym}
\usepackage{graphicx}
\usepackage[fleqn]{amsmath}
\usepackage{booktabs,tabularx, colortbl}
\usepackage{multirow}

\usepackage[T1]{fontenc}

\usepackage[utf8]{inputenc}

\usepackage{microtype}

\title{DU-VLG: Unifying Vision-and-Language Generation via Dual Sequence-to-Sequence Pre-training}

\author{Luyang Huang \quad
    {\bf Guocheng Niu} \quad
    {\bf Jiachen Liu} \quad
    {\bf Xinyan Xiao} \and
    {\bf Hua Wu} \\
    Baidu Inc., Beijing, China \\
  \texttt{\{huangluyang,niuguocheng,liujiachen,xiaoxinyan,} \\
  \texttt{wu\_hua\}@baidu.com} \\
  }

\begin{document}

\maketitle

\begin{abstract}

Due to the limitations of the model structure and pre-training objectives, existing vision-and-language generation models cannot utilize pair-wise images and text through bi-directional generation. In this paper, we propose \textbf{DU-VLG}, a framework which unifies vision-and-language generation as sequence generation problems. DU-VLG is trained with novel dual pre-training tasks: multi-modal denoising autoencoder tasks and modality translation tasks. To bridge the gap between image understanding and generation, we further design a novel commitment loss. We compare pre-training objectives on image captioning and text-to-image generation datasets. Results show that DU-VLG yields better performance than variants trained with uni-directional generation objectives or the variant without the commitment loss. On the image captioning task, our model reaches better performance than other pre-trained systems. On text-to-image generation datasets, our model achieves better or comparable results than previous state-of-the-art models. In addition, human judges further confirm that our model generates real and relevant images as well as faithful and informative captions.

\end{abstract}
\section{Introduction}


Pre-trained models for vision-and-language tasks have made remarkable progress recently~\cite{NEURIPS2019_c74d97b0, Su2020VL-BERT:,10.1007/978-3-030-58577-8_7}.  Existing pre-trained models either focus on text-to-image synthesis or image-to-text generation~\cite{DBLP:journals/corr/abs-2102-12092,pmlr-v139-cho21a}. These models are often pre-trained with image-text pairs which are aligned in semantics. However, due to the limitations of model structure, existing models cannot be adapted to each other. In addition, pre-training objectives are designed either for text generation conditioned on the image or image generation conditioned on the text, limiting the model to learn better semantic alignment from bi-directional generation~\cite{xu-etal-2021-e2e, DBLP:journals/corr/abs-2105-13290}.

\begin{figure}[t]
    \centering
    \includegraphics[width=\columnwidth,trim=0 0 0cm 0, clip]{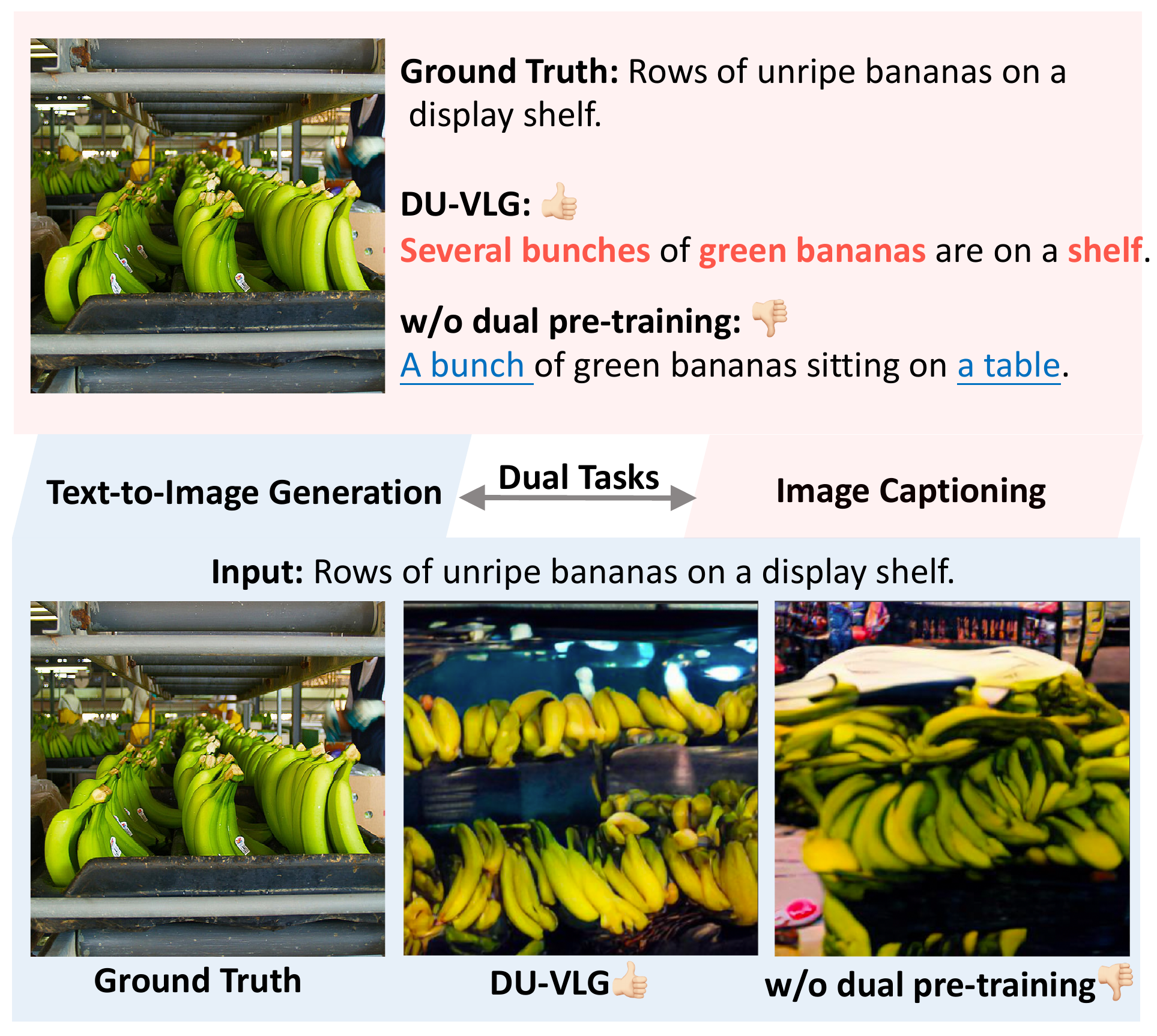}
    \captionof{figure}{An example from COCO dataset. For image captioning, our system generates informative captions, with key words highlighted in \textbf{\textcolor{red}{bold}}. Incorrect information is \underline{\textcolor{blue}{underlined}}. For text-to-image generation, our system synthesizes vivid images aligned with captions.
    }
    \label{fig:intro}
    \vspace{-2mm}
\end{figure}

We argue that \textit{image-to-text and text-to-image generation appear as dual tasks, which both require strong visual and textual representations aligned in the same semantic space.} Images and text descriptions are of different information quantity and density. The images often contain more information, but are with heavy redundancy, while text descriptions are semantically condensed, but may neglect details. Uni-directional generation paradigm may induce the model to amplify this property. Take Fig.\ref{fig:intro} as an example, the uni-directional model may fail in capturing details. Inspired by this observation, we propose to utilize bi-directional generation objectives to learn better generalization of image and text representations.





To this end, we present \textbf{DU-VLG}, a framework with \textbf{DU}al sequence-to-sequence pre-training for \textbf{V}ision-and-\textbf{L}anguage \textbf{G}eneration. Under the encoder-decoder Transformer framework, our model takes text and raw images as inputs and generate text and images autoregressively. Concretely, images are represented as continuous patch features in the encoder and discrete visual tokens in the decoder. With the hybrid image embedding schema, DU-VLG is able to unify vision-and-language generation in a single model.

In order to utilize dualities of image-text pairs, we further propose \textbf{two pairs of dual pre-training tasks}: multi-modal denoising autoencoder task and modality translation task. For the multi-modal denoising autoencoder task, our model takes image-text pairs with some image patches or words randomly masked as inputs and learns image-text alignment through reconstruction of the corrupted modality. For modality translation tasks, we form image captioning and text-to-image generation as dual pre-training tasks, which further enhance model ability of semantic alignment. Different from existing multi-modal pre-trained models, our model learns image-text alignment through bi-directional generation objectives.

Moreover, we propose \textbf{a novel commitment loss} to drive the model to acquire better image representation. Concretely, the commitment loss is designed to connect visual embeddings in the decoder to patch-based features in the encoder. In tandem with our model design, the commitment loss aims to unify image understanding and generation in a single model, which allows for better utilization of bi-directional generation objectives.

We conduct experiments on various vision-and-language generation tasks. We first study effects of dual pre-training tasks and the commitment loss. On both image captioning and text-to-image generation tasks, DU-VLG outperforms its variant without commitment loss or the variants that only learns uni-directional generation objectives. For image captioning, we achieve better BLEU-4 and CIDER than existing pre-trained models on COCO dataset~\cite{10.1007/978-3-319-10602-1_48}. For text-to-image generation, our model achieves better results than 
both Transformer-based and GAN-based methods on both COCO and CUB dataset~\cite{WelinderEtal2010}. Human judges confirm that our model generates captions and images with high-quality. Importantly, we test our model on a challenging vision-and-language generation task: visual commonsense reasoning~\cite{park2020visualcomet}. Results demonstrate that our model is able to handle challenging multi-modal generation tasks effectively.

The main contributions of DU-VLG are as follows:

\noindent $\bullet$ We unifies vision-and-language generation tasks with a single model, DU-VLG. With an encoder-decoder Transformer, DU-VLG is able to handle various vision-and-language generation tasks.

\noindent $\bullet$ DU-VLG is pre-trained with novel dual pre-training tasks, which utilizes dualities of image-text pairs. DU-VLG yields better or comparable results than existing state-of-the-art methods on three vision-and-language generation tasks.

\noindent $\bullet$ We further propose a new commitment loss, which aims to bridge the gap between image understanding and generation inner with our proposed dual paradigm. Experimental results show that the ability of dual tasks is further enhanced.

The rest of the paper is organized as follows. We describe our model in \S~\ref{sec:model} and introduce our proposed pre-training task and commitment loss in \S~\ref{sec:task}. Training details are presented in \S~\ref{sec:exp}. In \S~\ref{sec:results}, we discuss experimental results. Related work is listed in \S~\ref{sec:related} and we finally draw our conclusion in \S~\ref{sec:conclusion}.

\section{Model}
\label{sec:model}

\begin{figure*}[t]
    \centering
    \includegraphics[width=0.85\linewidth,trim=0 0cm 0 0, clip]{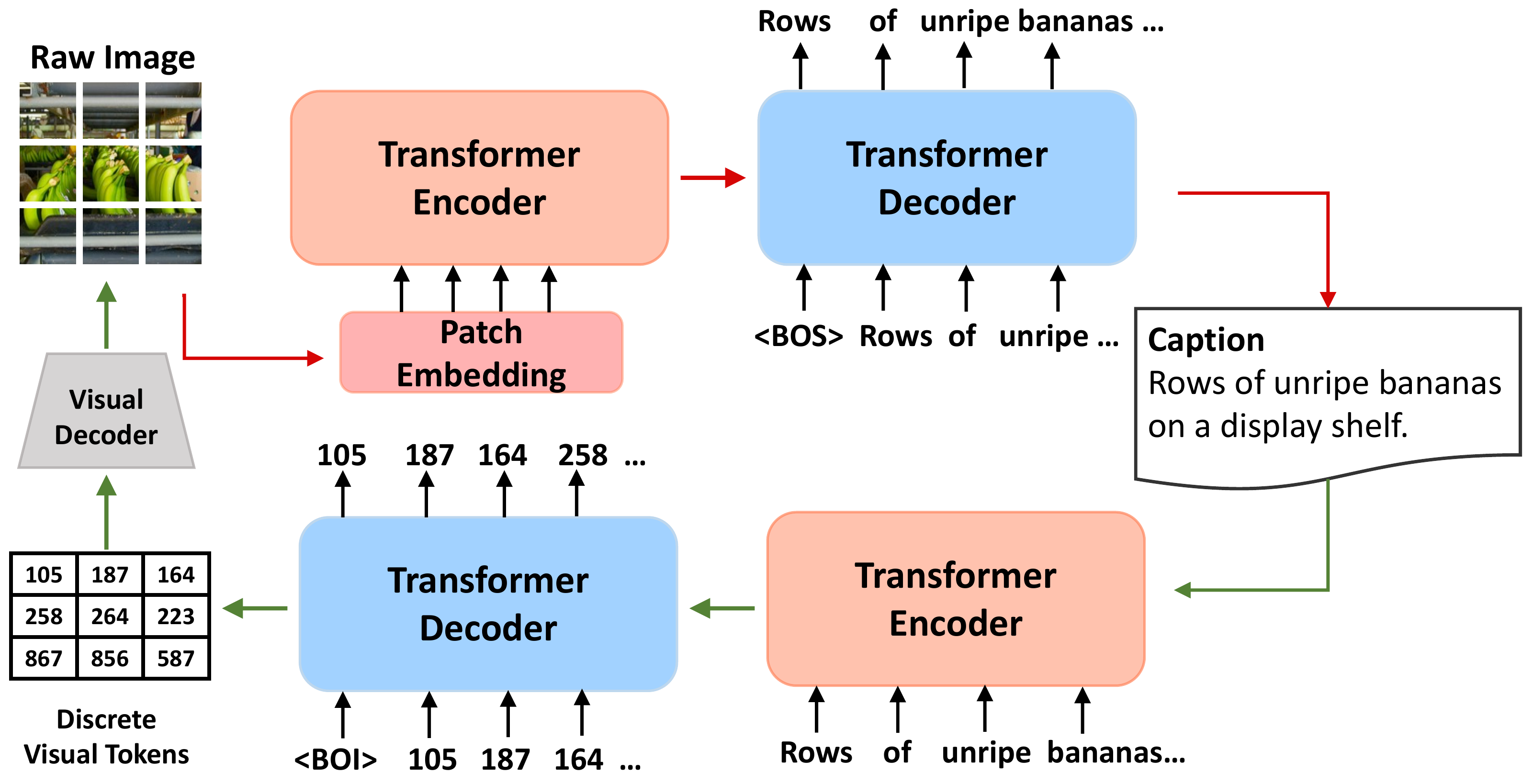}
    \captionof{figure}{ 
    An overview of DU-VLG. Our model is able to take images and text as inputs and generates images and text recurrently. In order to adapt image inputs to the Transformer-based model, we use a hybrid image embedding schema in encoder and decoder. The same color indicates that model parameters are shared for both images and text. The visual decoder weights are not used during training. The symmetric structure is designed for learning better representations from dual pre-training tasks.
    }
    \label{fig:model}
    \vspace{-2mm}
\end{figure*}

In this section, we describe our proposed model. Overall, our model design is mainly inspired by two observations: (1) sharing parameters that play the same role boosts model performance~\cite{pmlr-v80-xia18a} and (2) image understanding and generation require representing image features in different granularity~\cite{cho-etal-2020-x}. Hence, we use a standard Transformer with the encoder-decoder structure~\cite{NIPS2017_3f5ee243}, as illustrated in Fig.\ref{fig:model}. Our model takes images and text as inputs and treats image and text generation as sequence generation problems. Importantly, we propose to use a hybrid image embedding schema in the encoder and the decoder.


\subsection{Encoder}

In the encoder, images and text are first passed to embedding layers to obtain text embeddings $\mathbf{x}_{\rm text}$ and image embeddings $\mathbf{x}_{\rm image}$. For text embedding, we follow RoBERTa and tokenize inputs into BPEs~\cite{liu2020roberta}. Each BPE token is represented as the summation of word embedding and position embedding. Unlike text, Images are represented as pixels in a continuous semantic space. However, using pixels as image tokens results in a huge amount of computational cost since model needs to process long sequences. In order to maintain semantic information as well as reduce the computational cost, we split raw images into a grid of patches.

\smallskip
\noindent \textbf{Image Embedding for Encoder.} In the encoder, image inputs are flattened to a sequence of patches, with each patch represents the feature of $p$ $\times$ $p$ pixels. To obtain patch embedding, we pass input images to a trained Vision Transformer (ViT)~\cite{dosovitskiy2021an} and take hidden states of the last layer $\mathbf{x}_{\rm image}$ as image patch embeddings.

Image and text embeddings are then concatenated and fed into the encoder self-attention layers. If either image or text is missing in the input, we use a \texttt{[IMAGEPAD]} or \texttt{[TEXTPAD]} token as the placeholder.


\subsection{Decoder}

In the decoder, we use two embeddings: the text embedding which shares weights with the text embedding in the encoder and the image embedding which maps discrete visual tokens to embedding vectors. To enable autoregressive generation, we add \texttt{[BOI]} and \texttt{[EOI]} token to denote the start and the end of the image sequence. 

\smallskip
\noindent \textbf{Discrete Visual Tokens for Decoder.} In the decoder, the model generates a sequence of discrete visual tokens recurrently. During training, ground truth visual tokens are obtained by a Vector Quantised  Variational Autoencoder (VQ-VAE)~\cite{10.5555/3295222.3295378}. The VQ-VAE contains two modules, an image tokenizer and a visual decoder. The image tokenizer first extracts grid features from raw images and maps into discrete tokens $\mathbf{y}_{\rm image}$. The visual decoder reconstructs the original image from discrete visual tokens. The image tokenizer represents each $p$ $\times$ $p$ pixels as a visual token, with a vocabulary size of $|\mathcal{V}|$. Therefore, the number of decoder visual tokens is the same as the number of encoder patch tokens. We refer to the original paper for more details. Importantly, during testing, model first generates a sequence of image tokens recurrently and reconstruct the image with the visual decoder.

\section{Dual Pre-training Tasks and Pre-training Objectives}
\label{sec:task}

\begin{figure}[t]
    \centering
    \includegraphics[width=\columnwidth,trim=0 0 0cm 0, clip]{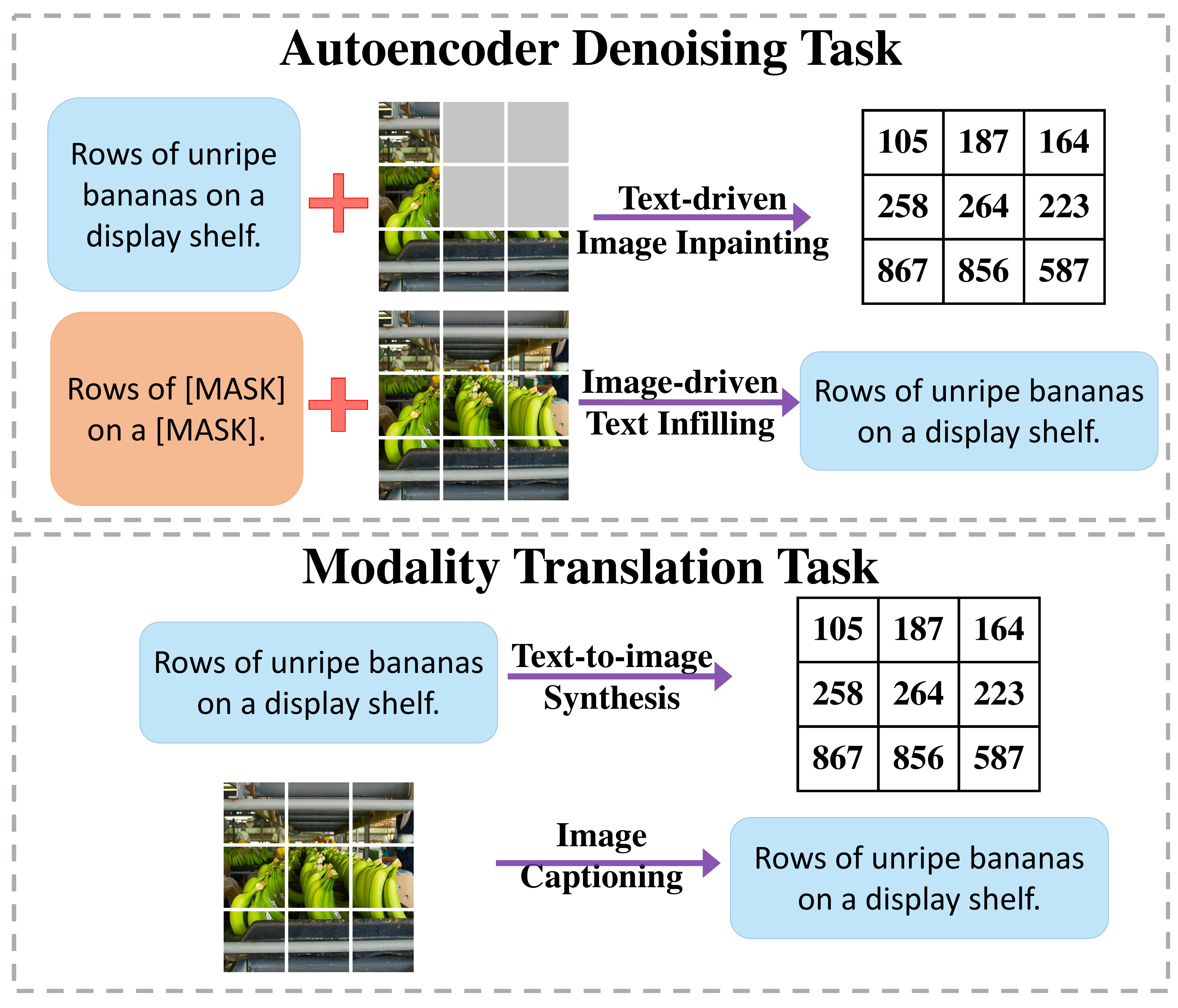}
    \captionof{figure}{ 
    An illustration of our proposed dual pre-training tasks. The model reconstructs the image or text conditioned on its visual and textual context.
    }
    \label{fig:task}
\end{figure}

Next, we introduce our pre-training method. Pre-training corpus consists of millions of aligned image-text pairs. In order to effectively learn vision-and-language understanding and generation, we propose dual pre-training tasks. Dual pre-training tasks drive the model to learn from reconstruction of the image or text description based on given context. We propose two pairs of pre-training tasks: (1) multi-modal denoising autoencoder task (\S~\ref{task:dae}) and (2) modality translation task (\S~\ref{task:mt}), as shown in Fig.\ref{fig:task}. In \S~\ref{task:commitment}, we formulate a commitment loss to connect image understanding and generation.


\subsection{Multi-modal Denoising Autoencoder Task}
\label{task:dae}

Given an image-text pair $(V,W)$ from the training set $D$, we first obtain image patch embeddings $\mathbf{x}_{\rm image}$ computed by ViT layers and attain text embeddings $\mathbf{x}_{\rm text}$. To encourage the model to learn cross-modal contextualized embeddings, we propose two dual tasks: 1) text-driven image inpainting task which aims to reconstruct the original image and 2) image-driven text infilling task which aims to reconstruct the original text.

\smallskip
\noindent \textbf{Text-Driven Image Inpainting.} Given image patch embeddings $\mathbf{x}_{\rm image}$, we replace 50 percent of image patches with the same umber of trainable \texttt{[MASK]} embeddings, producing masked image sequences $\mathbf{\Tilde{x}}_{\rm image}$. We use blockwise masking algorithm~\cite{DBLP:journals/corr/abs-2106-08254} to randomly select patches. Meanwhile, we feed the input image to the image tokenizer and produce a sequence of visual tokens $\mathbf{y}_{\rm image}$. The model is trained to reconstruct the image by optimizing negative log likelihood loss of the ground-truth visual tokens:

{\fontsize{10}{11}\selectfont
\begin{flalign}
\mathcal{L}^{\rm DAE}_{\rm image} &= - \sum_{(V, W) \in D} {\log{p(\mathbf{y}_{\rm image}\, | \mathbf{\Tilde{x}}_{\rm image}, \mathbf{x}_{\rm text})}}
\end{flalign} 
\label{eq:imageloss}
}  
\vspace{-3mm}

\smallskip
\noindent \textbf{Image-Driven Text Infilling.} Inspired by text infilling~\cite{lewis-etal-2020-bart}, we randomly sample a number of text spans from a Poisson distribution ($\lambda = 3$) and replace with a single \texttt{[MASK]}. Different from text infilling, we randomly mask 50 percent of tokens since we additionally include image as visual context. The model is trained to optimize negative log likelihood loss of original text tokens:

{\fontsize{10}{11}\selectfont
\begin{flalign}
\mathcal{L}^{\rm DAE}_{\rm text} &= - \sum_{(V, W) \in D} {\log{p(\mathbf{x}_{\rm text}\, | \mathbf{\Tilde{x}}_{\rm text}, \mathbf{x}_{\rm image})}}
\end{flalign} 
\label{eq:textloss}
}  
\vspace{-3mm}

where $\mathbf{\Tilde{x}}_{\rm text}$ represents the corrupted text sequence.

\subsection{Modality Translation Task}
\label{task:mt}

In addition to the denoising autoencoder task, we further enhance the model with the modality translation task. The modality translation task drives the model to learn mapping from a modality to the other. Given an image-text pair, we form the modality translation task as two dual tasks: 1) image captioning and 2) text-to-image synthesis.

\smallskip
\noindent \textbf{Image Captioning.} Given an image as input, model first produces image patch embeddings $\mathbf{x}_{\rm image}$ from ViT and encodes image features with encoder self-attentions. The decoder is trained to generate text based on image features. The loss function can be defined as:

{\fontsize{10}{11}\selectfont
\begin{flalign}
\mathcal{L}^{\rm MT}_{\rm text} &= - \sum_{(V, W) \in D} {\log{p(\mathbf{x}_{\rm text}\, | \mathbf{x}_{\rm image})}}
\end{flalign} 
\label{eq:textloss_2}
}  
\vspace{-3mm}

\smallskip
\noindent \textbf{Text-to-Image Synthesis.} Given a visual description as input, model encodes the input with the encoder and the decoder generates discrete visual tokens $\textbf{y}_{\rm image}$ recurrently. During training, the ground truth visual tokens are computed by the image tokenizer. The loss function can be defined as:

{\fontsize{10}{11}\selectfont
\begin{flalign}
\mathcal{L}^{\rm MT}_{\rm image} &= - \sum_{(V, W) \in D} {\log{p(\mathbf{y}_{\rm image}\, | \mathbf{x}_{\rm text})}}
\end{flalign} 
\label{eq:imageloss_2}
}  
\vspace{-3mm}

\subsection{Connecting Image Embedding between Encoder and Decoder.}
\label{task:commitment}

In the encoder-decoder structure, text embedding is often shared among the encoder, the decoder and the token generation layer~\cite{paulus2018a}. This allows the model to learn better syntactic and semantic information. For image embedding, since we use a hybrid embedding schema in the encoder and the decoder, we propose a commitment loss to connect image understanding and generation during training. Intuitively, decoder visual token embeddings $\mathbf{y}_{\rm image}$ should commit to corresponding patch embeddings $\mathbf{x}_{\rm image}$ in encoder. Therefore, the commitment loss uses a square loss to connect the encoder and the decoder:

{\fontsize{10}{11}\selectfont
\begin{flalign}
\mathcal{L}_{\rm com} &= - \sum_{(V) \in D} {\parallel {\rm sg}[\mathbf{x}_{\rm image}] - \mathbf{y}_{\rm image} \parallel^{2}}
\end{flalign} 
\label{eq:commitment}
}  
\vspace{-3mm}

where ${\rm sg}$ means stopgradient operator which is identity at forward computation but has zero partial derivatives at backward computation. The commitment loss is applied to the text-driven image inpainting objective and the text-to-image synthesis objective.

During training, for each instance, we randomly select a couple of objectives from denoising autoencoder and modality translation. We set probability of denoising autoencoder as 0.6 for all experiments. Therefore, for each batch, the pre-training loss is a combination of three losses:

{\fontsize{10}{11}\selectfont
\begin{flalign}
\mathcal{L}_{\rm total} &= \mathcal{L}_{\rm text} + \alpha \mathcal{L}_{\rm image} \\
\mathcal{L}_{\rm image} &= \mathcal{L}^{DAE}_{\rm image} + \mathcal{L}^{MT}_{\rm image} + \beta \mathcal{L}_{\rm com} \\ 
\mathcal{L}_{\rm text} &= \mathcal{L}^{DAE}_{\rm text} + \mathcal{L}^{MT}_{\rm text} 
\end{flalign} 
\label{eq:ptloss}
}  
\vspace{-3mm}

where $\alpha$ and $\beta$ are hyperparameters to control the scale of image loss and commitment loss.

\section{Experimental Setup}
\label{sec:exp}
\subsection{Pre-training}

\smallskip
\noindent \textbf{Pre-training Corpus.} We train our model on four existing datasets that consist of image-text pairs. Our pre-training datasets include 1) Common Objects in Context (COCO)~\cite{10.1007/978-3-319-10602-1_48}, 2) Conceptual Captions (CC)~\cite{sharma-etal-2018-conceptual}, 3) SBU Captioned Photo (SBU)~\cite{Ordonez:2011:im2text} and 4) Visual Genome (VG)~\cite{Krishna2016VisualGC}. For Visual Genome dataset, since captions are collected for image regions, we use image regions and captions as pairs. We additionally filter captions which are fewer than five words. We end up with a collection of about 5 million image-text pairs.

\smallskip
\noindent \textbf{Implementation Detail.} We report results on two model sizes: 1) a base version with 6 layers for the encoder and decoder and 2) a large version with 12 layers for the encoder and decoder. For each model size, we report results with two different input image resolutions: 224 $\times$ 224 and 384 $\times$ 384. Following ViT, we use a patch size of $p=16$ for all the experiments. For VQ-VAE, we take the off-the-shelf VQ-GAN~\cite{Esser_2021_CVPR}, which is a variant of VQ-VAE. The VQ-GAN maps each $16 \times 16$ pixels as a discrete visual token, with a vocabulary size of $|\mathcal{V}| = 16384$.

For base and large model, we use \texttt{ViT-base} and \texttt{ViT-large} with a patch size of $p=16$ to extract image patch embeddings. ViT weights are set frozen during pre-training. Since image sequences are longer than text sequences, we set $\alpha = 0.05$ and $\beta=1$ for all experiments. For model optimization, we utilize Adam optimizer with a gradient clipping of 1.0 and a batch size equivalent of 1024.

\subsection{Fine-tuning on Downstream Tasks}

In order to evaluate model capability of vision-and-language generation tasks, we test on three downstream tasks: 1) text-to-image generation, 2) image captioning and 3) visual commonsense reasoning. Here we mainly introduce evaluation metrics. For additional fine-tuning details, we refer to the appendices.

\smallskip
\noindent \textbf{Text-to-Image Generation.} We experiment with two popular text-to-image generation datasets: the Caltech-UCSD Birds 200 dataset (CUB) and Common Objects in Context dataset (COCO). 

The CUB dataset contains 200 bird categories with 11,788 images. Each image has ten text descriptions. We follow the standard split which uses 150 categories with 8,855 images for training and the remaining 50 categories with 2,933 images for testing. The COCO dataset contains 82,784 images for training and 40,505 for testing. Each image has five text descriptions.

We fine-tune on the pre-trained model with a learning rate of 1e-4 for 300 epoches on both datasets. Similar to ~\newcite{DBLP:journals/corr/abs-2102-12092}, we sample 16 images per caption with nucleus sampling strategy~\cite{Holtzman2020The}.
During testing, we first sample 16 images per caption and rerank the generated images with a CLIP model~\cite{pmlr-v139-radford21a}. The CLIP model selects the best image based on its correlation with the text description. 

We include two widely used evaluation metrics: 1) Inception Score (IS)~\cite{NIPS2016_8a3363ab} and 2) Fréchet Inception Distance (FID)~\cite{NIPS2017_8a1d6947}. The IS score computes the KL-divergence between the conditional class distribution and the marginal class distribution obtained by a pre-trained Inception v3 model~\cite{7780677}. The FID computes the Fréchet distance between ground-truth images and generated images based on the features obtained by the Incaption v3 model. Higher IS scores and lower FID scores denote that images synthesized by the model are of better quality. Previous work~\cite{Li_2019_CVPR} reports that the IS score fails in evaluating the quality of images on COCO dataset. Hence, we do not report the IS score on COCO dataset. For fair comparison, we resize our model outputs to $256 \times 256$ and calculate FID and IS scores.

\smallskip
\noindent \textbf{Image Captioning.} For image captioning, we test our model on COCO dataset. We report four metrics based on word overlapping on COCO dataset: 1) BLEU-4~\cite{papineni-etal-2002-bleu}, 2) METEOR~\cite{lavie-agarwal-2007-meteor}, 3) CIDEr~\cite{Vedantam_2015_CVPR} and 4) SPICE~\cite{johnson-etal-2020-spice}. 

For COCO dataset, we follow the Karparthy split~\cite{Karpathy_2015_CVPR} which has 113,287, 5000 and 5000 images for training, validation and test. Each image has 5 human-written captions. During inference, we generate a caption for each image and evaluate against five references. 

We fine-tune on COCO dataset with a learning rate of 3e-5. 
Vision Transformer layers are trainable during fine-tuning. Following ~\newcite{10.1007/978-3-030-58577-8_8}, we add object labels detected by the object detection model as additional text inputs. We find object labels improve CIDER and BLEU scores for at least 1 point and 0.3 points. During testing, we use beam search with a beam size of 5.

\smallskip
\noindent \textbf{Visual Commonsense Reasoning.} Besides image captioning and text-to-image generation, which only requires model to encode one modality, we further test our model on a more challenging dataset, VisualCOMET~\cite{park2020visualcomet}. VisualCOMET is a visual commonsense reasoning task which provides the model with an image and the event that happens at present. The model is required to infer what may happen next, before and the people's intents at present. VisualCOMET requires the model to jointly comprehend image and text and generate reasonable inference. Similar to image captioning, we use BLEU-2, METEOR and CIDEr as metrics.

\section{Results}
\label{sec:results}

In this section, we start with comparing our proposed pre-training objectives in \S~\ref{results:sec1}. We then conduct automatic evaluation on three vision-and-language generation tasks (\S~\ref{results:sec2}) and further report human evaluation on both caption and synthesized image quality (\S~\ref{results:sec2}). Finally, we investigate inference speed of our proposed model (\S~\ref{results:sec3}).

\begin{table}[t]
\centering
\fontsize{10}{11}\selectfont
\setlength{\tabcolsep}{0.5mm}
\begin{tabular}{@{}lcccc@{}}
\toprule
 Image -> Text & \multicolumn{4}{c}{COCO Caption} \\
\textbf{System} & \textbf{BLEU-4} & \textbf{CIDER} & \textbf{METEOR} & \textbf{SPICE} \\
 \midrule
 \textsc{DU-VLG}$_{\text{B}-224}$ & \textbf{38.8} & \textbf{124.8} & \textbf{29.2} & \textbf{22.0}  \\
 w/o $L_{\rm image}$ & 36.9 & 118.8 & 28.4 & 20.5 \\
 w/o $L_{\rm text}$ & 35.2 & 112.8 & 27.4 & 19.6 \\ 
 w/o $L_{\rm com}$ & 38.4 & 123.1 & 28.8 & 21.7 \\ 
 \toprule 
Text -> Image & \multicolumn{2}{c}{CUB} & COCO & \\
 \textbf{System} & \textbf{IS}$\uparrow$ & \textbf{FID}$\downarrow$ & \textbf{FID}$\downarrow$ &   \\
\textsc{DU-VLG}$_{\text{B}-224}$ & \textbf{5.14} & \textbf{23.78} & \textbf{26.82} & \\
 w/o $L_{\rm image}$  & 4.84 & 25.28 & 36.59 & \\
 w/o $L_{\rm text}$ & 5.03 & 24.68 & 29.64 & \\
 w/o $L_{\rm com}$ & 5.08 & 24.44 & 27.92 &  \\ 
\bottomrule
\end{tabular}
\vspace{-2mm}
\caption{Ablation study on pre-training tasks and objectives. The best result per metric per dataset is \textbf{bolded}. \textsc{DU-VLG}$_{\text{B}-224}$ yields significantly higher scores than other comparisons with approximate randomization test ($p < 0.0005$).
}
\label{tab:ablation_study}
\vspace{-1mm}
\end{table}

\subsection{Comparing Pre-training Objectives}
\label{results:sec1}

\smallskip
\noindent \textbf{Comparisons.} We first investigate whether our proposed dual pre-training tasks and commitment loss improve generation quality. We fine-tune on two downstream tasks: image captioning and text-to-image generation. We report our base model with an input image resolution of $224 \times 224$ ( \textsc{DU-VLG}$_{\text{B}-224}$). We compare our base model with three variants: 1) the model trained without text-driven image inpainting and text-to-image synthesis tasks (w/o $L_{\rm image}$), 2) the model trained without image-driven text infilling and image captioning tasks (w/o $L_{\rm text}$) and 3) the model trained without commitment loss (w/o $L_{\rm com}$).

\smallskip
\noindent \textbf{Results.} As displayed in Tab.\ref{tab:ablation_study}, \textit{our model with dual pre-training tasks performs the best on both image captioning and text-to-image generation tasks}. This demonstrates the benefit of dual pre-training tasks and the commitment loss. For image captioning, comparing with the variant without image generation objectives, our model with dual pre-training tasks significantly improves automatic metrics, which indicates that image generation objectives can boost visual understanding. For text-to-image generation, our model yields better FID and IS scores than the variant without text generation objectives on both CUB and COCO dataset. This demonstrates that using text generation objectives can guide better semantic interpretation of text content.

Moreover, our model outperforms the variant trained without the commitment loss on two downstream tasks. This further illustrates 
that the commitment loss improves model performance on both image understanding and generation.

\subsection{Automatic Evaluation}
\label{results:sec2}

\smallskip
\noindent \textbf{Comparisons.} We then compare our model with other vision-and-language models. For image captioning, we include state-of-the-art vision-and-language pre-trained models: (1) object-semantics aligned pre-training (\textsc{OSCAR})~\cite{10.1007/978-3-030-58577-8_8}, (2) unified modal understanding and generation pre-training (\textsc{UNIMO})~\cite{li-etal-2021-unimo}, (3) improving visual representations for vision-and-language pre-training (\textsc{VinVL})~\cite{Zhang_2021_CVPR} and (4) end-to-end vision-and-language pre-training (\textsc{E2E-VLP})~\cite{xu-etal-2021-e2e}. For \textsc{OSCAR} and \textsc{VINVL}, we report their results with cross-entropy optimization for fair comparison. 

For text-to-image generation, we include four Transformer-based models: (1) \textsc{X-LXMERT}, which has 228 million parameters and is trained on 9 million image-text pairs, (2) \textsc{DALLE}, which has 12 billion parameters and is trained on 250 million text-image pairs 
~\cite{DBLP:journals/corr/abs-2102-12092}, (3) \textsc{Cogview}, which has 4 billion parameters and is trained on 30 million data~\cite{DBLP:journals/corr/abs-2105-13290} and (4) \textsc{NUWA}, which has 870 million parameters and is trained on a mixture of text-image pairs and text-video pairs~\cite{DBLP:journals/corr/abs-2111-12417}. We further compare our model with three traditional methods based on generative adversarial network (GAN): (1) \textsc{DM-GAN}~\cite{Zhu_2019_CVPR}, (2) \textsc{DF-GAN}~\cite{DBLP:journals/corr/abs-2008-05865} and (3) \textsc{XMC-GAN}~\cite{zhang2021cross}.

For visual commonsense reasoning, we include Vision-Language Transformer (\textsc{V-L Transformer})~\cite{park2020visualcomet} as a baseline, which fuses region-based visual features into a pre-trained GPT-2~\cite{radford2019language}. 

\smallskip
\noindent \textbf{Results.} For image captioning, our model achieves better scores than both end-to-end method and two-stage methods. In Tab.\ref{tab:caption}, DU-VLG outperforms previous state-of-the-art pre-trained model \textsc{VINVL}, e.g., improving BLEU-4 and CIDEr by more than 1 and 3 points. 

Moreover, for text-to-image generation tasks, our model achieves state-of-the-art IS and FID on CUB dataset, as displayed in Tab.\ref{tab:text2image}, outperforming traditional GAN-based methods. Compared with Transformer-based methods, our model yields better or comparable FID scores on COCO datasets. It is worth to note that our models are with fewer parameters and less training data compared with \textsc{DALLE}, \textsc{Cogview} and \textsc{NUWA}. This demonstrates the effectiveness of our proposed framework.

In addition, we study the effect of different input image resolutions. We compare two different resolutions of the input images: $224 \times 224$ and $384 \times 384$. In Tab.\ref{tab:caption} and Tab.\ref{tab:text2image}, we find higher resolution as inputs leads to better results on both image-to-text and text-to-image generation tasks. This observation remarks the importance of fine-grained image representation.

\begin{table}[t]
\centering
\fontsize{10}{11}\selectfont
\setlength{\tabcolsep}{0.5mm}
\begin{tabular}{@{}lcccc@{}}
\toprule
 Image -> Text & \multicolumn{4}{c}{CoCo Caption} \\
\textbf{System} & \textbf{BLEU-4} & \textbf{CIDER} & \textbf{METEOR} & \textbf{SPICE}     \\
\midrule
 \textsc{OSCAR}$_{\text{B}}$ & 36.5 & 123.7 & 30.7 & 23.5  \\
 \textsc{UNIMO}$_{\text{B}}$ & 38.8 & 124.4 & 29.8 & 22.1 \\
 \textsc{VINVL}$_{\text{B}}$ & 38.2 & 129.3 & 30.3 & 23.6 \\
 \textsc{E2E-VLP} & 36.2 & 117.3 & -- & -- \\
 \midrule
 \textsc{DU-VLG}$_{\text{B}-224}$ & 38.8 & 124.8 & 29.2 & 22.0 \\
 \textsc{DU-VLG}$_{\text{B}-384}$ & \textbf{40.0} & \textbf{133.0} & 30.2 & \textbf{23.8} \\
\midrule 
 \textsc{OSCAR}$_{\text{L}}$ & 37.4 & 127.8 & 30.7 & 23.5 \\
 \textsc{UNIMO}$_{\text{L}}$ & 39.6 & 127.7 & 29.5 & 22.4 \\
 \textsc{VINVL}$_{\text{L}}$ & 38.5 & 130.8 & 30.4 & 23.4 \\
 \midrule 
  \textsc{DU-VLG}$_{\text{L}-224}$ & 39.2 & 128.1 & 29.8 & 22.8  \\
 \textsc{DU-VLG}$_{\text{L}-384}$ & \textbf{40.1} & \textbf{135.8} & \textbf{30.8} & \textbf{23.9} \\

\bottomrule
\end{tabular}
\caption{Automatic evaluation on Image Captioning datasets. We report our model and comparisons with two model sizes: the base version (${\text{B}}$) and the large version (${\text{L}}$) and two input image resolution: $224 \times 224$ and $384 \times 384$. Our base and large models have comparable number of parameters compared to other comparisons. The best metric of each model size is \textbf{bolded}.
}
\label{tab:caption}
\vspace{-1mm}
\end{table}

We then evaluate our model on a more challenging vision-and-language task, visual commonsense reasoning. As shown in Tab.\ref{tab:vcr}, our model significantly outperforms \textsc{V-L Transformer}, which is fine-tuned based on a language model, GPT-2. This demonstrates that our model is able to jointly comprehend image and text inputs and generate informative inference.

\subsection{Human Evaluation}
\label{results:sec3}

\begin{table}[t]
\centering
\fontsize{10}{11}\selectfont
\setlength{\tabcolsep}{2mm}
\begin{tabular}{@{}lccc@{}}
\toprule
 Text -> Image & \multicolumn{2}{c}{CUB} & \multicolumn{1}{c}{COCO} \\
\textbf{System} & \textbf{IS}$\uparrow$ & \textbf{FID}$\downarrow$ & \textbf{FID}$\downarrow$   \\\midrule
 \textsc{DM-GAN} & 4.75 & 16.09 & 32.64 \\
 \textsc{DF-GAN} & 5.10 & 14.81 & 21.42 \\
 \textsc{XMC-GAN} & -- & -- & \textbf{9.33} \\
 \textsc{X-LXMERT} & -- & -- & 37.40 \\
 \textsc{DALLE} & -- & -- & 27.50 \\
 \textsc{COGVIEW} & -- & -- & 26.00 \\
 \textsc{NUWA} & -- & -- & 12.90 \\
\midrule
 \textsc{DU-VLG}$_{\text{B-224}}$ & 5.14 & 23.78 & 26.82 \\
  \textsc{DU-VLG}$_{\text{B-384}}$ & 5.26 & 14.60 & 22.41 \\
  \midrule
  \textsc{DU-VLG}$_{\text{L-224}}$ & 5.18 &  21.50 & 23.25  \\
 \textsc{DU-VLG}$_{\text{L-384}}$ & \textbf{5.28} & \textbf{14.15} & 14.48  \\

\bottomrule
\end{tabular}
\caption{Automatic evaluation on Text-to-Image Generation datasets. For fair comparison, we resize generated images to $256 \times 256$ pixels before calculating IS and FID scores.
}
\label{tab:text2image}
\vspace{-1mm}
\end{table}

\begin{table}[t]
\centering
\fontsize{10}{11}\selectfont
\setlength{\tabcolsep}{0.5mm}
\begin{tabular}{@{}lccc@{}}
\toprule
 & \multicolumn{3}{c}{VisualCOMET} \\
\textbf{System} & \textbf{BLEU-2} & \textbf{CIDER} & \textbf{METEOR} \\
\midrule 
\textsc{V-L Transformer} & 13.5 & 18.2 & 11.5  \\
 \midrule
 \textsc{DU-VLG}$_{\text{B}-384}$ & 21.5 & 36.6 & 25.6  \\
 \textsc{DU-VLG}$_{\text{L}-384}$ & \textbf{23.9} & \textbf{41.9} & \textbf{27.1}  \\

\bottomrule
\end{tabular}
\caption{Automatic evaluation on visual commonsense reasoning. Our model generates informative inference compared to the baseline.
}
\label{tab:vcr}
\vspace{-2mm}
\end{table}

\begin{figure}[t]
    \centering
    \includegraphics[width=\columnwidth,trim=0 0 0cm 0, clip]{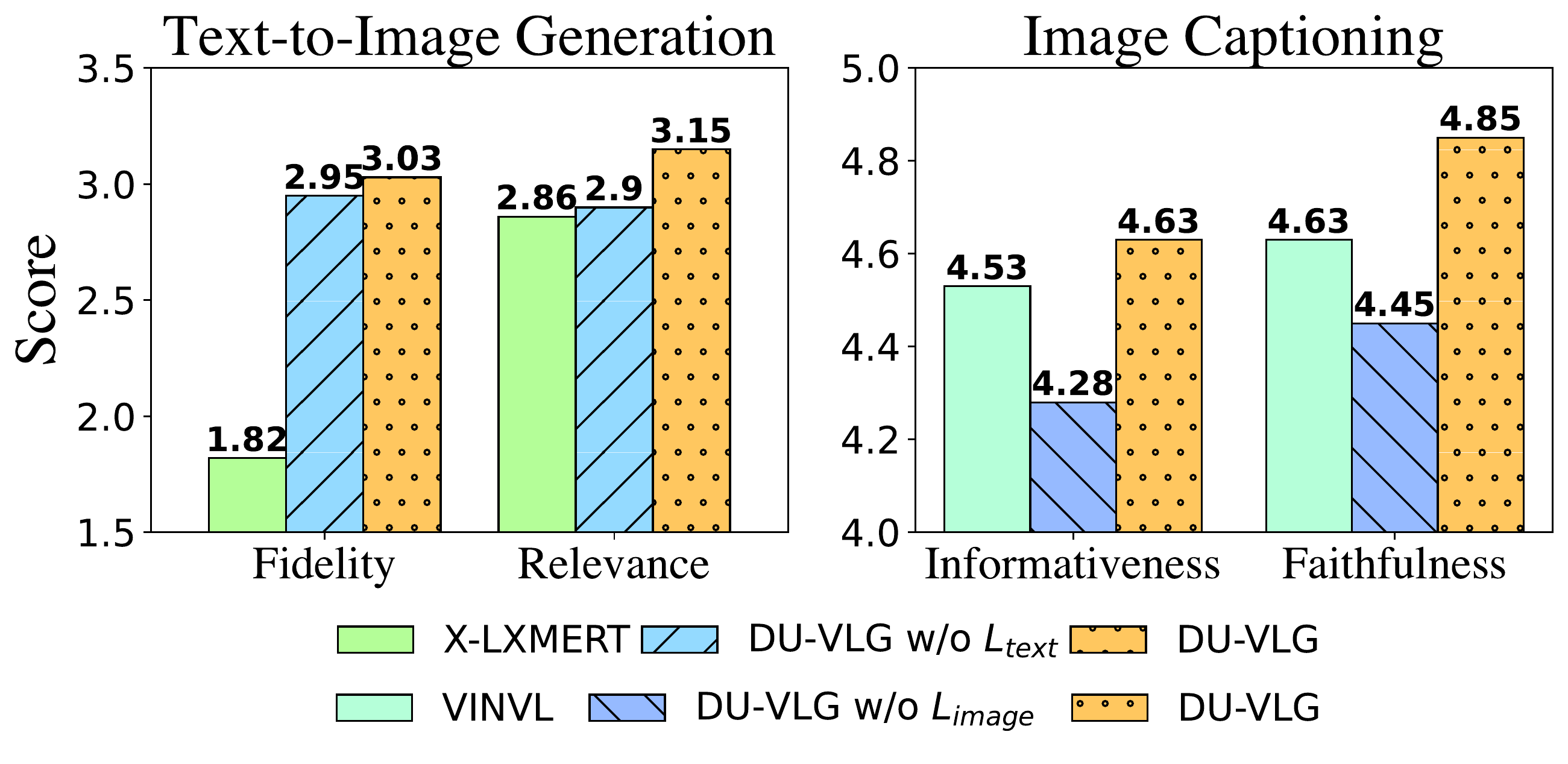}
    \captionof{figure}{Human evaluation on COCO dataset.\textsc{DU-VLG} yields significantly higher scores than other systems on fidelity, relavance, informativeness and faithfulness ($p < 0.05$).
    }
    \label{fig:humaneval}
    \vspace{-1mm}
\end{figure}

We conduct human evaluation to analyze generation quality of images and text. For both image captioning and text-to-image generation, we select 100 samples from COCO test set and hire three annotators to rate captions and images. For image captioning, we include three systems: (1) best performed pre-trained model \textsc{VINVL} (2) our model that removes dual pre-training \textsc{DU-VLG} w/o $L_{\rm image}$ and (3) our best performed model \textsc{DU-VLG}. For text-to-image generation, we compare three models: (1) Transformer-based model pre-trained on about 9 million data \textsc{X-LXMERT}, (2) our model trained without text generation objectives \textsc{DU-VLG} w/o $L_{\rm text}$ and (3) \textsc{DU-VLG}. For our model, we use the large version with the input image resolution of $384 \times 384$.

For image captioning, human judges are asked to rate on two aspects: \textbf{informativeness}---whether the caption covers important objects from the image and \textbf{faithfulness}---whether the caption correctly describes the image. For text-to-image generation, we consider two aspects: \textbf{fidelity}---whether the image is realistic and \textbf{relevance}---whether the image matches with the caption. All aspects are rated on a Likert scale from 1 (poor) to 5 (good).

\smallskip
\noindent \textbf{Results.} From Fig.\ref{fig:humaneval}, we find our \textsc{DU-VLG} model obtains better scores in relevance, fidelity, informativeness and faithfulness than the variant that removes dual pre-training tasks. This confirms our claim that bi-directional generation objectives improve semantic alignment between images and text. Meanwhile, compared with well-performed model \textsc{VINVL} and \textsc{X-LXMERT}, our model yields better scores on four aspects. This implies that our model generates more informative captions committed to the input images and synthesizes more realistic images aligned with the captions compared to the state-of-the-art pre-trained models. 

Interestingly, image captioning models yield higher scores than text-to-image generation models, closer to 5 (perfect). After inspection, we find that our model yields near-perfect captions compared to human written ones, while the generated images sometimes fail in synthesizing details. For example, the shape of a banana may be distorted, limiting the fidelity of the image.

\subsection{Inference Efficiency}
\label{results:sec4}

Next, we compare the inference speed and the number of model parameters with existing models. For image captioning, we compare our model with two best performed pre-trained models: the base version of UNIMO and VINVL. For text-to-image generation, we compare with two transformer-based large models DALLE and Cogview. For our model, we report the base version. We test speed on COCO test set with one 32GB NVIDIA TESLA V100. We include the visual decoder when calculating the inference speed.

In Tab.\ref{tab:speed_comparison}, we find our model is roughly 7$\times$ faster than two-stage methods on image captioning. This is mainly because extracting image features with ViT is much faster than object detection. Importantly, our model has comparable parameters compared with UNIMO and VINVL.

For text-to-image generation, our model is roughly 400$\times$ faster than large model Cogview and has only 5 percent of parameters. This further confirms the importance of dual pre-training tasks.

\begin{table}[t]
\centering
\fontsize{11}{11}\selectfont
\setlength{\tabcolsep}{2mm}
\begin{tabular}{@{}lcc@{}}
\toprule
\textbf{System} & \textbf{Time(s)}  & \textbf{\# Param. (M)}   \\
\midrule
\multicolumn{2}{@{}l}{\bf Image Captioning} & \\
\textsc{UNIMO}$_{\text{B}}$ & 0.88$+$0.12 & 172 \\
\textsc{VINVL}$_{\text{B}}$ & 0.90$+$0.12 & 187 \\
\textsc{DU-VLG}$_{\text{B-224}}$ & 0.14 & 228 \\
\midrule
\multicolumn{2}{@{}l}{\bf Text-to-Image Generation}  & \\
\textsc{DALLE}  & -- & 12,000 \\
\textsc{Cogview} &  300 & 4,000 \\
\textsc{DU-VLG}$_{\text{B-224}}$ & 0.76 & 228 \\
\bottomrule
\end{tabular}
\caption{Comparing inference speed (time) and number of parameters (\# Param.) on different tasks. For two-stage methods \textsc{UNIMO} and \textsc{VINVL}, we report image feature extraction and beam search time respectively.
}
\label{tab:speed_comparison}
\vspace{-3mm}
\end{table}

\section{Related Work}
\label{sec:related}

\smallskip
\noindent \textbf{Vision-and-Language Pre-training for Image-to-Text Generation Tasks.} Transformer backbones have achieved great success in language pre-training~\cite{devlin-etal-2019-bert,lewis-etal-2020-bart,liu2020roberta}. In order to adapt Transformers to multi-modal pre-training, previous work mainly focuses on (1) better image features
 and (2) designing pre-training tasks~\cite{NEURIPS2019_c74d97b0, DBLP:journals/corr/abs-1908-03557}. 
 To obtain high-quality image features, Image region features extracted from an object detection model are widely adopted in multi-modal pre-training~\cite{Zhou_Palangi_Zhang_Hu_Corso_Gao_2020,10.1007/978-3-030-58577-8_8,Zhang_2021_CVPR}. ~\newcite{pmlr-v139-kim21k} points out that two-stage method is time-consuming and the trained object detector may fail in the unlabeled domain~\cite{jiang2021decoupled}. To that end, 
~\newcite{DBLP:journals/corr/abs-2004-00849} 
feeds raw images to convolutional backbones such as ResNets~\cite{7780459} and takes its outputs as image features. ~\newcite{pmlr-v139-kim21k} uses linear projection to obtain patch-based image features. However, currently, end-to-end image feature extraction methods cannot yield comparable results compared to two-stage methods on image captioning.
 
To learn image-text alignment, masked token prediction, which masks a portion of text or image tokens and predicts masked positions conditioned on the context, is widely used as the pre-training task~\cite{DBLP:journals/corr/abs-2003-01473}.  ~\newcite{10.1007/978-3-030-58577-8_8} designs image-text matching task, which  predicts whether the image and the text are paired or not. ~\newcite{li-etal-2021-unimo} proposes special self-attention masks to unify text understanding and generation. ~\newcite{xu-etal-2021-e2e} includes image captioning and object detection as pre-training objectives to enhance the decoder. However, current methods for generation tasks are limited to text generation and are struggled to learn fine-grained image-text alignment.

In this paper, we introduce a hybrid image embedding schema to connect image understanding and generation, which unifies image and text generation via sequence-to-sequence pre-training. Concretely, we enhance image-text alignment with novel dual pre-training tasks. Our model outperforms state-of-the-art pre-trained systems on image captioning.

\smallskip
\noindent \textbf{Vision-and-Language Pre-training for Text-to-Image Generation Tasks.} To generate images autoregressively, images are represented as discrete tokens.
X-LXMERT~\cite{cho-etal-2020-x} partitions image grid features into clusters and obtains visual tokens via neareast-neighbor search. However, X-LXMERT needs to train an image generator from scratch to synthesize images from visual tokens, which accumulates errors during training. ~\newcite{DBLP:journals/corr/abs-2105-13290,DBLP:journals/corr/abs-2102-12092} use discrete visual tokens from a trained vector-quantised variational autoencoder (VQ-VAE)~\cite{10.5555/3295222.3295378} for text-to-image generation. However, their models consist of billions of parameters and require a huge corpus to pre-train (more than 100 million image-text pairs). In this paper, we present a relative small model (about 200M parameters), with better generation quality on COCO dataset. In particular, we offer a detailed analysis on the inference speed and the model size in the appendices.



\section{Conclusion}
\label{sec:conclusion}

We presented a novel framework, DU-VLG, which unifies vision-and-language generation tasks with an encoder-decoder Transformer. We propose to use a hybrid image embedding schema in the encoder and decoder. In addition, we pre-train the model with novel dual pre-training tasks, along with a new commitment loss, to guide better image and text understanding and generation. Experiments show that our proposed dual pre-training objectives significantly improve performance on three vision-and-language generation tasks. Human evaluation further confirms that our model with dual pre-training tasks improves generation quality on image captioning and text-to-image generation.

\section{Acknowledgments}

This work was supported by the National Key Research and Development Project of China (No. 2018AAA0101900)

\section{Ethics Statement}

Large models that are pre-trained on heterogeneous data can be potentially harmful to marginalized populations. Along with the improved controllability, we also recognize that our system might be misused to create offensive or fabricated content. We therefore advocate cautious usage in real-world deployment.

\bibliography{anthology,custom}
\bibliographystyle{acl_natbib}

\appendix

\clearpage

\section{Additional Evaluation}

\begin{figure*}[t]
    \centering
    \includegraphics[width=0.95\linewidth,trim=0 0cm 0 0, clip]{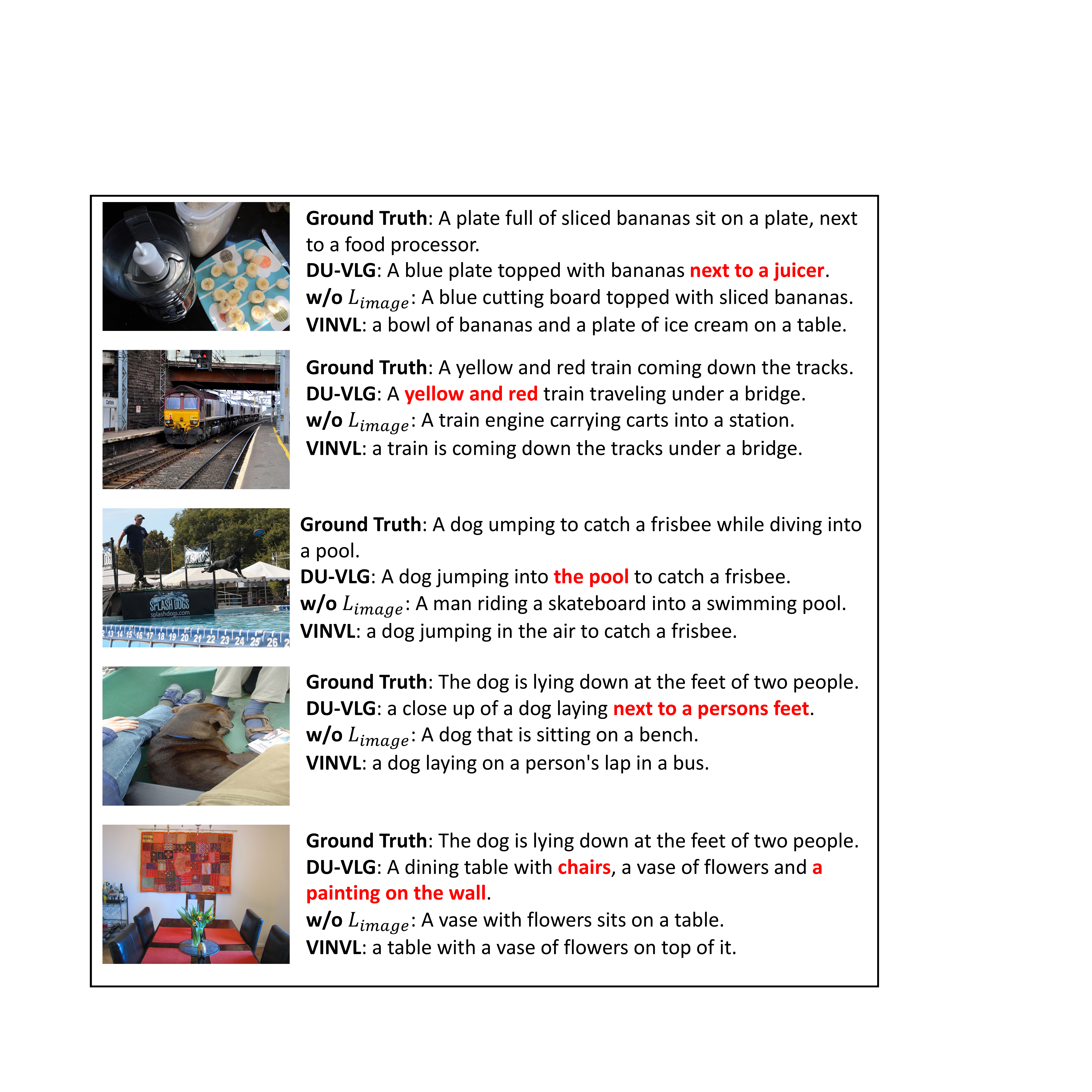}
    \captionof{figure}{ 
    Samples on image captioning from COCO dataset. DU-VLG generates faithful and informative captions, highlighted in \textbf{\textcolor{red}{red}}.
    }
    \label{fig:image2text}
\end{figure*}

\begin{figure*}[t]
    \centering
    \includegraphics[width=0.9\linewidth,trim=0 2cm 0 0, clip]{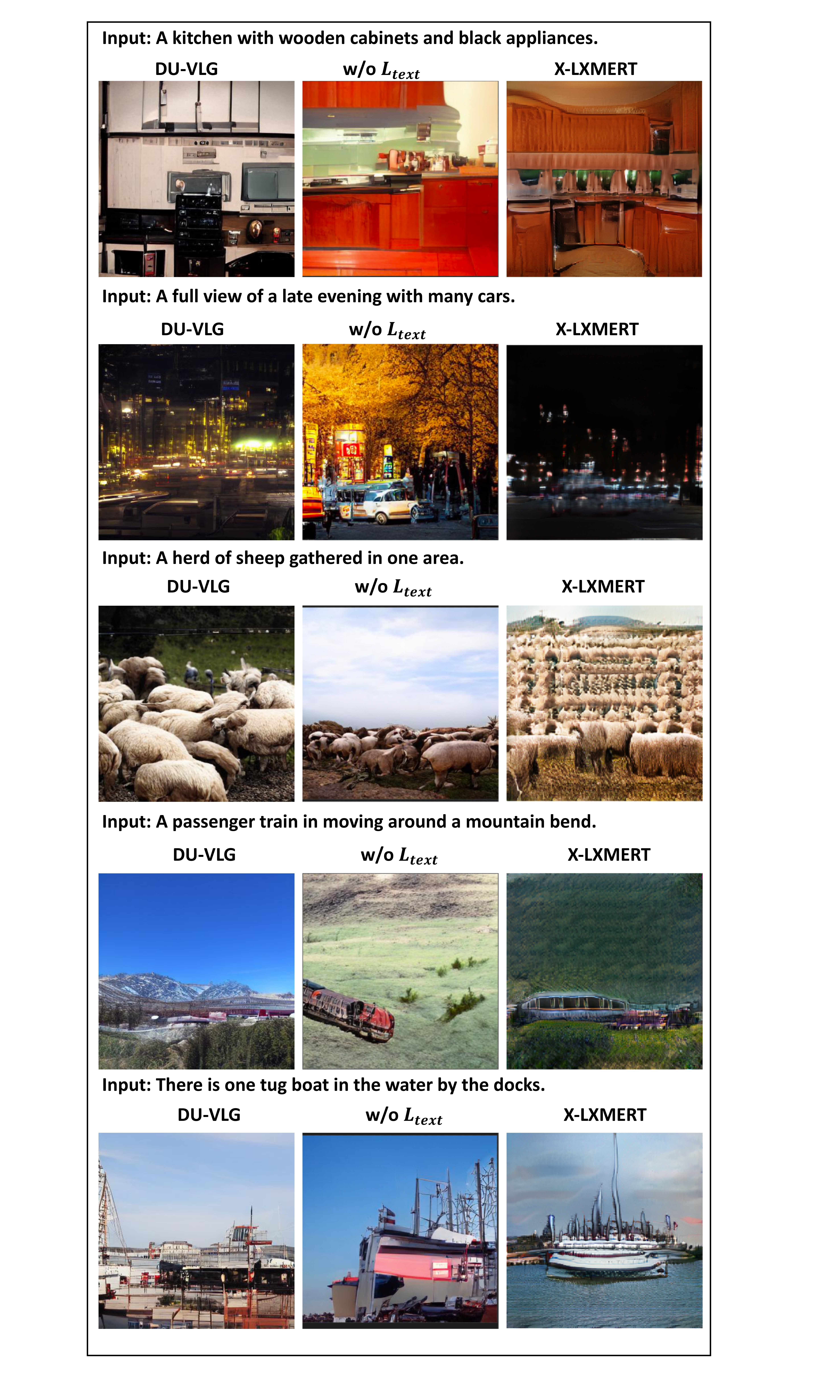}
    \captionof{figure}{ 
    Samples on text-to-image generation from COCO dataset. DU-VLG generates vivid and relevant images.
    }
    \label{fig:text2image}
\end{figure*}

We include 5 examples on COCO dataset for image captioning and text-to-image generation tasks. In Fig.\ref{fig:image2text} and Fig.\ref{fig:text2image}, we find that DU-VLG generates captions and images of high quality.

\section{Human Evaluation Guideline}

In human evaluation, each annotator is presented with 100 model generated images and 100 model generated captions from 3 systems (in random order). For text-to-image generation, the human judges are asked to evaluate on fidelity and informativeness on a scale of 1 to 5 (1 being good and 5 being poor). Here are descriptions of two aspects:

$\bullet$ \textbf{Fidelity}: Whether the image is realistic and looks like a real photo.

$\bullet$ \textbf{Relevance}: Whether the image provides necessary content coverage from the text description.

For image captioning, the human annotators are asked to evaluate on faithfulness and informativeness on a scale of 1 to 5 (1 being good and 5 being poor). Here are detailed descriptions of two aspects:

$\bullet$ \textbf{Faithfulness}: Whether the caption correctly describes main objects in the image.

$\bullet$ \textbf{Informativeness}: Whether the caption covers enough information from the image.

The definition of four aspects can be found in Tab.\ref{tab:guideline}.

\newpage
\begin{table}
	\fontsize{10}{11}\selectfont
    \begin{tabular}{lp{65mm}}
         \toprule
         \multicolumn{2}{c}{\textbf{Image Captioning}} \\
         \midrule
         \multicolumn{2}{c}{\textbf{Informativeness:}} \\
         \midrule
         \rowcolor{lightgray!30}
          1 & Not relevant to the image. \\
          \rowcolor{lightgray!30}
          3 & Relevant, but misses the main objects of the image. \\
          \rowcolor{lightgray!30}
          5 &  Successfully captures the main point of the image. \\
          \midrule
         \multicolumn{2}{c}{\textbf{Faithfulness:}} \\
         \midrule
         \rowcolor{lightgray!30}
         1 & The caption is full of fabricated content.  \\
         \rowcolor{lightgray!30}
         3 & The caption is overall relevant to the image, but contains some fake details. \\
         \rowcolor{lightgray!30}
         5 & The caption matches with the image. \\
          \midrule
         \multicolumn{2}{c}{\textbf{Text-to-Image Generation}} \\
         \midrule
         \multicolumn{2}{c}{\textbf{Fidelity:}} \\
         \midrule
         \rowcolor{lightgray!30}
          1 & The image is unreal, distorted or blurred. \\
          \rowcolor{lightgray!30}
          3 & The image is overall realistic, but some details are blurred or distorted. \\
          \rowcolor{lightgray!30}
          5 & The image is vivid and looks like a real photo. \\
          \midrule
         \multicolumn{2}{c}{\textbf{Relavance:}} \\
         \midrule
         \rowcolor{lightgray!30}
         1 & The image does not match with the caption.  \\
         \rowcolor{lightgray!30}
         3 & The image is related to the caption, but some details are hallucinated. \\
         \rowcolor{lightgray!30}
         5 & The image clearly reflects the caption. \\

         \bottomrule
    \end{tabular}
    \caption{The definition of four aspects in human evaluation.}
    \label{tab:guideline}
\end{table}

~\\

\newpage

\end{document}